\documentclass{article}
\usepackage{spconf,amsmath,graphicx}
\usepackage{breqn}
\usepackage{csquotes}
\usepackage{url}
\graphicspath{
{./figures/}
}

\title{Show, Translate and Tell}
\makeatletter
\def\@name{ \emph{Dheeraj Peri, \enskip Shagan Sah, \enskip Raymond Ptucha}}
\makeatother

\address{Rochester Institute of Technology\\
        Rochester NY, USA}
\begin{document}
%\ninept
%
\maketitle
%
%%%%%%%%%%%%%%%%%%%%%%%%%%%%%%%%%%%%%%%%%%%%% ABSTRACT
\begin{abstract}

Humans have an incredible ability to process and understand information from multiple sources such as images, video, text, and speech. Recent success of deep neural networks has enabled us to develop algorithms which give machines the ability to understand and interpret this information. There is a need to both broaden their applicability and develop methods which correlate visual information along with semantic content. We propose a unified model which jointly trains on images and captions, and learns to generate new captions given either an image or a caption query. We evaluate our model on three different tasks namely cross-modal retrieval, image captioning, and sentence paraphrasing. Our model gains insight into cross-modal vector embeddings, generalizes well on multiple tasks and is competitive to state of the art methods on retrieval.

% and achieves state-of-the-art performance on retrieval.
% Our model is  and is able to generalize well on multiple tasks.
\end{abstract}

%%%%%%%%%%%%%%%%%%%%%%%%%%%%%%%%%%%%%%%%%%%%% BODY TEXT
\section{Introduction}

Image and text understanding has seen significant progress with the proliferation of convolutional and recurrent neural networks. These neural networks have accomplished outstanding results when applied to individual tasks such as image captioning, cross-modal retrieval and visual-question answering. In each of these tasks, domain transformation is learnt to transfer information between images and text. Off-the-shelf pre-trained networks \cite{He2015} have been used widely to extract features which represent the objects and their relationships in an image. 

Multi-task learning \cite{devlin2018bert} \cite{zheng2018same} has recently been applied to natural language processing. By training on multiple tasks jointly, a model learns abstract representations that are task agnostic. This approach can effectively be used as pre-training and is shown to improve many natural language processing tasks \cite{devlin2018bert}. We investigate this approach and apply it to vision and language tasks namely cross-modal retrieval, image captioning and sentence paraphrasing. Our model learns generalized latent representations of image and text.  This approach reduces both inference time and memory requirements when compared to task-specific models. 

\begin{figure*}
  \label{stt_model}
  \centering
  \includegraphics[width=1.0\linewidth]{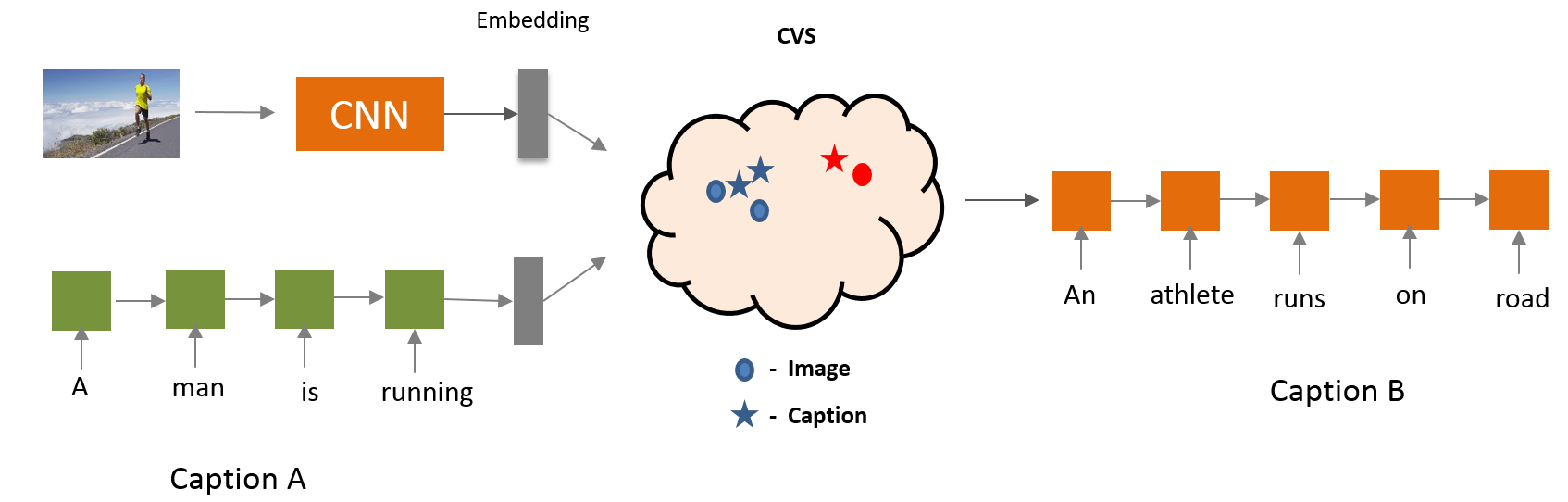}
  \caption{Show, Translate and Tell model. During training, an image and caption A are projected into CVS and the embeddings are decoded into caption B, which is semantically similar to caption A. Inside the CVS, the blue points indicate correlated image and caption pairs and the red points are uncorrelated with the blue ones.}
  \label{fig:stt_main_fig}
\end{figure*}

The main contributions of this work include:
\begin{itemize}
    \item We propose a unified model that jointly trains on images and captions and learns to generate new captions given either an image or text as query.
    \item We apply this multi-task model to the three different tasks namely cross-modal retrieval, image captioning and sentence paraphrasing.
    \item We leverage an attention mechanism in this multi-task model and demonstrate improved performance on these tasks.
    \item We open source our implementation at \url{https://github.com/peri044/STT}.
    % which improves the generalization performance.
\end{itemize}
%%%%%%%%%%%%%%%%%%%%%%%%%%%%%%%%%%%%%%%%%%%%%
\section{Related Work} \label{sec:literature_review}
Recent progress in deep learning has enabled significant advancements in understanding the relationships between visual and language entities. Most of the works focus on extracting advanced deep features and trying to map them to a specific tasks such as image captioning, phrase localization, and cross-modal retrieval. 
    
% Koch et al \cite{koch2015siamese} introduced siamese networks to learn similarities between characters using contrastive loss and achieved superior performance on one-shot image recognition tasks.
Aviv et al. \cite{eisenschtat2017linking} used two way neural networks to optimize euclidean loss between images and text in a common embedding space. Vendrov et al. \cite{vendrov2015order} proposed to use an order-violation penalty in margin-based ranking loss \cite{karpathy2015deep} to enforce constraint on the order in which embeddings are learned. In particular, they only use the absolute value of image and text embeddings and use margin-based loss to optimize the model. Faghri et al. \cite{faghri2017vse++} proved hard negative mining can be useful and showed significant improvements on cross modal retrieval problems. Wehrmann et al. \cite{brm} proposed to use convolutional text encoders and perform convolutions over characters as opposed to words. They use an embedding matrix for characters and show significant reduction in the number of parameters of the model. You et al. \cite{cse} proposed to use a local loss along with a global loss to train the image embeddings. Yan et al. \cite{huang2018learning} proposed to use a multi-label CNN to predict semantic concepts in the image. They use an LSTM network as a sentence encoder to represent sentences and apply margin based ranking loss to bring images and sentences into a common embedding space. Lee et al. \cite{lee2018stacked} proposed a novel attention mechanism to align image regions with the individual words in a sentence. They compute the attention scores as a similarity metric and optimize the margin based ranking loss. 

In this work, we propose a unified model which learns general purpose representations for both images and text which can be applied in the context of multi-task learning. 
%%%%%%%%%%%%%%%%%%%%%%%%%%%%%%%%%%%%%%%%%%%%%%%%%%%%%%
\section{Show, Translate and Tell}

Show, Translate and Tell (STT) is a unified model, which projects images and captions into a common embedding space, and also learns to decode these embeddings into meaningful representations. This approach offers an interesting and insightful way to interpret the semantics of these embeddings.

\subsection{Method} 
Figure \ref{fig:stt_main_fig} describes the high-level architecture. 
During training, images and captions are encoded using Convolutional Neural Networks (CNN) and Recurrent Neural Networks (RNNs) respectively. These features are projected into a Common Vector Space (CVS) by using fully connected layers. To align similar concepts of images and text together and map dissimilar concepts far apart in this CVS, we follow \cite{karpathy2015deep} and enforce a margin based ranking loss to optimize the model parameters. This loss ensures that the model learns the semantic relationships between images and captions. The margin-based ranking loss is given by (1).
 
\begin{equation}
\begin{array}{r}
L_{rank} = \sum\limits_{m}\sum\limits_{k} max(0, \alpha + s(i,c_k) - s(i,c)) \\  \\+ \sum\limits_{k}\sum\limits_{m} max(0, \alpha + s(c,i_m) - s(c,i))
\end{array}
\end{equation}

where $s(i,c)$ denotes the cosine similarity between correlated image-caption pair and $s(i, c_k)$ denotes the cosine similarity between uncorrelated image-caption pair. $i$ and $c$ denote the image and caption embeddings obtained from the image and text encoders. This loss forces the similarity of correlated pairs to be higher than uncorrelated pairs by a minimum margin $\alpha$. 

\noindent
\textbf{Decoding Image embeddings -}
To ensure that the image embeddings are closer to the captions in CVS, 
% To encourage image embeddings to be closer to caption embeddings,
we decode the image embeddings using a second RNN (on the right of the Figure 1) into semantically similar captions. We use cross entropy loss at each time step of the RNN, given by (\ref{ic_loss}).

\begin{equation}
\label{ic_loss}
\begin{array}{r}
L_{IC} = -\sum\limits_{t=1}^{N} log P(w_t|I; \theta)
\end{array}
\end{equation}

where ${P(w_t)}$ is the probability of predicting word ${w_t}$ at timestep $t$, $I$ is the image embeddings from CVS and $\theta$ denotes the parameters of the RNN and the image encoder.

\noindent
\textbf{Decoding Sentence embeddings -}
In order to ensure sentence embeddings have semantic meaning, we also decode these embeddings using the same RNN (on the right of the Figure 1), into semantically similar captions. We use cross entropy loss at each timestep of the RNN, given by (\ref{sp_loss}).

\begin{equation}
\label{sp_loss}
\begin{array}{r}
L_{SP} = -\sum\limits_{t=1}^{N} log P(w_t|S; \theta)
\end{array}
\end{equation}

where ${P(w_t)}$ is the probability of predicting word ${w_t}$ at timestep $t$, $S$ is the input sentence embeddings from CVS and $\theta$ denotes the parameters of the RNN and the sentence encoder. The weights of this RNN are shared during the decoding of image and sentence embeddings.

Combining the above components, the STT model is jointly trained by optimizing the overall objective function in (\ref{total_loss}).

\begin{equation}
\label{total_loss}
\begin{array}{r}
L = \lambda_1L_{rank} + \lambda_2L_{IC} + \lambda_3L_{SP}
\end{array}
\end{equation}

where $\lambda_1$, $\lambda_2$ and $\lambda_3$ are scalar weights which regulate the importance of individual loss components.

\subsection{STT model with Attention}

In order to align fine-grain information in the STT model, we follow \cite{lee2018stacked} to incorporate attention between image regions and individual words in the sentence. Object proposals are extracted using Faster R-CNN \cite{girshick2015fast} and top $N$ region proposals are passed through a Resnet-152 CNN for feature extraction. Captions are encoded by an LSTM network and the outputs at individual time steps are collected. The importance of each word over the $N$ regions is calculated by cosine distance and the similarity metric $s(i,c)$ is computed as the aggregate of all the word vectors and regions. During training with attention, we use the average of $N$ region level embeddings as an input to the decoder during image captioning. 

\section{Implementation}
   
We use TensorFlow \cite{abadi2016tensorflow} package throughout our experiments. We resize the images into $256\times256$ and extracted random crops of $224\times224$ for data augmentation. We use Resnet-152 CNN \cite{he2016deep} as a feature extractor for images and 1-layer LSTM \cite{hochreiter1997long} network to encode and decode sentences. We use one fully connected layer for image and sentence embeddings. We pre-compute the ResNet-152  \cite{he2016deep} features and train the image embedding, RNN encoder and decoder modules for 15 epochs. The learning rate is set to 0.0002 and Adam optimizer is used to optimize the parameters of the model. The margin parameter and batch size was set to 0.2 and 128 respectively. For attention models, we follow \cite{lee2018stacked} and choose $N$ to be 36.

\section{Experiments}

We perform extensive experiments to evaluate the proposed model. We choose standard datasets, MSCOCO \cite{lin2014microsoft} and Flickr30K for training and evaluation.  We use 113,287 images from MSCOCO and 29,783 images from Flickr30K for training, 5000 and 1000 images for evaluation. During training, a single sample constitutes one image and two sentence paraphrases. Since each image in \cite{lin2014microsoft} and \cite{flickr30k} is annotated with five captions, we form a total of 20 permutations of paraphrases for each image in the dataset. We report recall ($R@K$) as the evaluation metric for retrieval experiments. $R@K$ is the percentage of query samples in which the ground-truth sentences belong to the top $K$ retrieved sentences. For MSCOCO, we report the average of 5-fold cross validation results on 5000 test images. We use BLEU \cite{papineni2002bleu} and METEOR \cite{banerjee2005meteor} scores for captioning and paraphrasing experiments.

\subsection{Cross Modal Retrieval}

Cross Modal Retrieval is the task of retrieving similar samples given an input query sample. The input query can either be an image or a caption. The images and captions in the test set are passed through image and caption encoders in Figure \ref{stt_model} to extract their corresponding embeddings.

\begin{table}[!ht]
\centering
\caption{Results of Cross Modal Retrieval on MSCOCO.}
\begin{tabular}{c|c|c|c|c}
\hline
Model & \multicolumn{2}{c}{Sentence Retrieval} & \multicolumn{2}{c}{Image Retrieval} \\
& R@1 & R@10 & R@1 & R@10 \\ \hline
UVS \cite{uvs} &43.4 & 85.8 & 31.0 & 79.9 \\
Order \cite{vendrov2015order} &46.7 & 88.9 & 37.9 & 85.9 \\
Aviv \cite{eisenschtat2017linking} &55.8 & - & 39.7 & - \\
BRM \cite{brm} &55.1 & 93.9 & 41.2 & 89.2 \\
CSE \cite{cse}  &56.3 & 91.5 & 45.7 & 90.6 \\
VSE++ \cite{faghri2017vse++}& 58.3 & 93.3 & 43.6  & 87.8 \\
STT  & \textbf{55.1} & \textbf{92.1} & \textbf{41.0} & \textbf{86.0} \\
STT with att & \textbf{64.9}  & \textbf{96.8} & \textbf{49.8} & \textbf{91.6} \\
FBB \cite{engilberge2018finding} & 69.8 & 96.6 & 55.9 & 94.0 \\
SCO \cite{huang2018learning} & 69.9 & 97.5 & 56.7 & 94.8 \\
SCAN \cite{lee2018stacked} & 72.7 & 98.4 & 55.8 & 94.8 \\
\hline
\end{tabular}
\label{result_mscoco}
\end{table}   

\begin{table}[!ht]
\centering
\caption{Results of Cross Modal Retrieval on Flickr30k.}
\begin{tabular}{c|c|c|c|c}
\hline
Model & \multicolumn{2}{c}{Sentence Retrieval} & \multicolumn{2}{c}{Image Retrieval} \\
& R@1 & R@10 & R@1  & R@10 \\ \hline
UVS \cite{uvs}  &29.8 & 70.5 & 22  & 59.3 \\
BRM \cite{brm} &36.0 & - & 31.0 & - \\
VSE++ \cite{faghri2017vse++} & 43.7 & 82.1 & 32.3 & 87.8 \\
CSE \cite{cse} &44.6 & 83.8 & 36.9  & 79.6 \\
Order \cite{vendrov2015order} &46.7 &  88.9 & 37.9  & 85.9 \\
STT  & \textbf{38.4}  & \textbf{77.5} & \textbf{27.1} & \textbf{68.2} \\
FBB \cite{engilberge2018finding} & 46.5 & 82.2 & 34.9 & 73.5 \\ 
SCO \cite{huang2018learning} & 55.5 & 89.3 & 41.1 & 80.1 \\
STT with att & \textbf{59.2}  & \textbf{91.0} & \textbf{40.7} & \textbf{79.0} \\
SCAN \cite{lee2018stacked} & 67.9 & 94.4 & 43.9 & 82.8 \\
 \hline
\end{tabular}
\label{result_flickr}
\end{table} 

Tables \ref{result_mscoco} and \ref{result_flickr} show cross modal retrieval results of STT model on MSCOCO and Flickr30K respectively. From Table \ref{result_mscoco} and \ref{result_flickr}, it is clear that STT model performs well on cross modal retrieval task. STT with attention performs better than all other models on Flickr30K \cite{flickr30k}, except for SCAN \cite{lee2018stacked}. While SCAN \cite{lee2018stacked} does great at cross modal retrieval, our model can additionally do captioning and paraphrasing.

% Although, it doesn't perform better than the other models, the embeddings of STT are more generalizable and can be used to perform diverse tasks like captioning and paraphrasing.

\subsection{Image Captioning}

Images are passed through the encoders and the corresponding embeddings are passed through an RNN decoder in Figure \ref{fig:stt_main_fig} to generate new captions. Since the model learned to map semantics from different modalities, these embeddings can generate meaningful sentence representations.
%  including words and concepts not directly attributed with a supervised training sample.
Table \ref{im_capt} shows the results of image captioning on a test set of 5000 and 1000 images respectively. We use the same test set for retrieval and captioning tasks. 

\begin{table}[htbp]
\centering
\caption{Results of Image Captioning using STT model.}
\label{im_capt}
\begin{tabular}{|r|r|r|r|r|r|}
    \hline
Dataset & B@1 & B@2 & B@3 & B@4 & METEOR \\ \hline
\multicolumn{6}{|c|}{w/o Attention} \\ \hline
\multicolumn{1}{|l|}{MSCOCO} & 0.683 & 0.506 & 0.362 & 0.259 & 0.236  \\ \hline
\multicolumn{1}{|l|}{Flickr30k} & 0.513 & 0.330 & 0.204 & 0.129 & 0.178 \\ \hline
\multicolumn{6}{|c|}{with Attention} \\ \hline
\multicolumn{1}{|l|}{MSCOCO} & 0.706 & 0.530 & 0.385 & 0.279 & 0.246 \\ \hline
\multicolumn{1}{|l|}{Flickr30k} & 0.611 & 0.427 & 0.293 & 0.203 & 0.193 \\ \hline
\end{tabular}
\end{table}

\subsection{Sentence Paraphrasing}

Similar to captioning, we can use the RNN encoder and decoder to generate paraphrases. The input sentence is passed through an RNN encoder and the generated embeddings are passed through an RNN decoder in Figure \ref{stt_model} to generate a caption with similar meaning.
% Since the model learned to map semantics from different modalities, these embeddings would generate meaningful sentence representations.
The test set for MSCOCO has 5000 images which results in a total of 25000 ground truth captions. Flickr30k has a total of 5000 ground truth captions (1000 images). To the best of our knowledge, we are the first ones to show sentence paraphrasing in a multi-task setting between vision and language. Table
 \ref{sent_par} shows the results of sentence paraphrasing on MSCOCO and Flickr30K.

\begin{table}[htbp]
\centering
\caption{Results of Sentence Paraphrasing using STT model.}
\label{sent_par}
\begin{tabular}{|r|r|r|r|r|r|}
    \hline
Dataset & B@1 & B@2 & B@3 & B@4 & METEOR \\ \hline
\multicolumn{6}{|c|}{w/o Attention} \\ \hline
\multicolumn{1}{|l|}{MSCOCO} & 0.744 & 0.578 & 0.435 & 0.324 & 0.275  \\ \hline
\multicolumn{1}{|l|}{Flickr30k} & 0.569 & 0.394 & 0.262 & 0.176 & 0.217 \\ \hline
\multicolumn{6}{|c|}{with Attention} \\ \hline
\multicolumn{1}{|l|}{MSCOCO} & 0.747 & 0.581 & 0.436 & 0.326 & 0.272 \\ \hline
\multicolumn{1}{|l|}{Flickr30k} & 0.673 & 0.493 & 0.353 & 0.252 & 0.221 \\ \hline
\end{tabular}
\end{table}

% \subsection{Visual Results}
\begin{figure}
  \label{stt_model}
  \centering
  \includegraphics[width=1.0\linewidth]{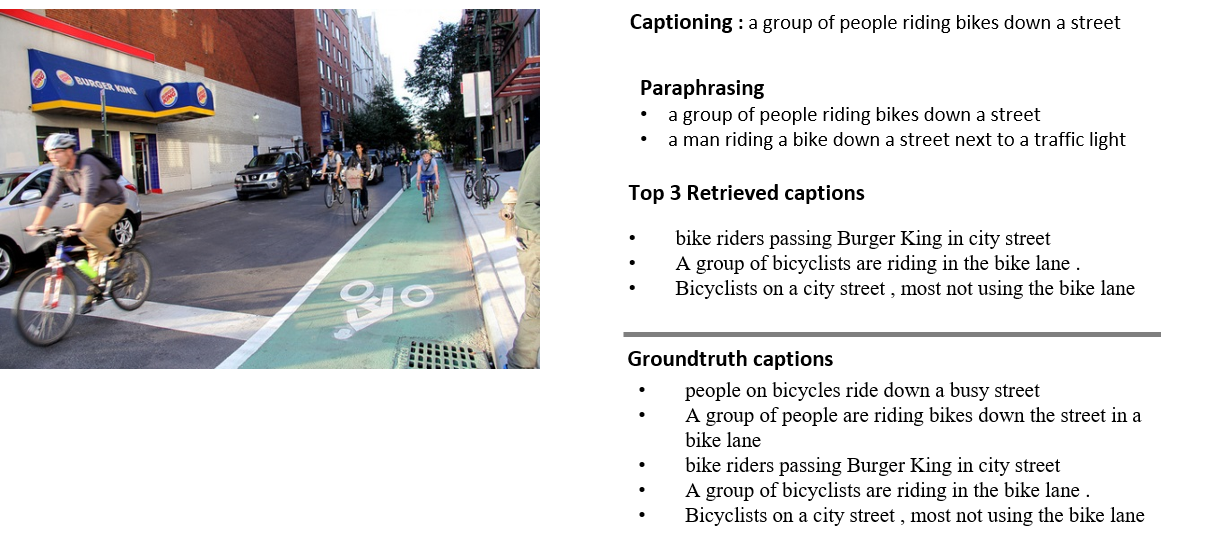}
  \caption{Sample result of STT model on MSCOCO dataset.}
  \label{fig:stt_coco}
\end{figure}
\begin{figure}
  \label{stt_model}
  \centering
  \includegraphics[width=1.0\linewidth]{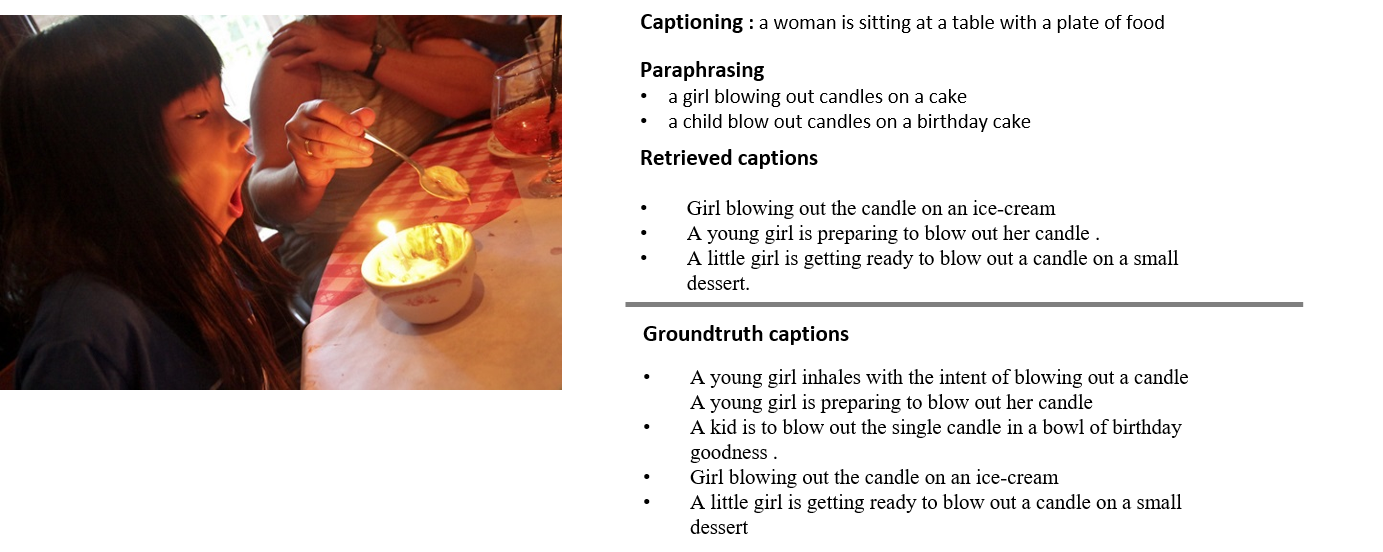}
  \caption{Sample result of STT model with Attention on MSCOCO dataset.}
  \label{fig:stt_coco_2}
\end{figure}
\begin{figure}
  \label{stt_model}
  \centering
  \includegraphics[width=1.0\linewidth]{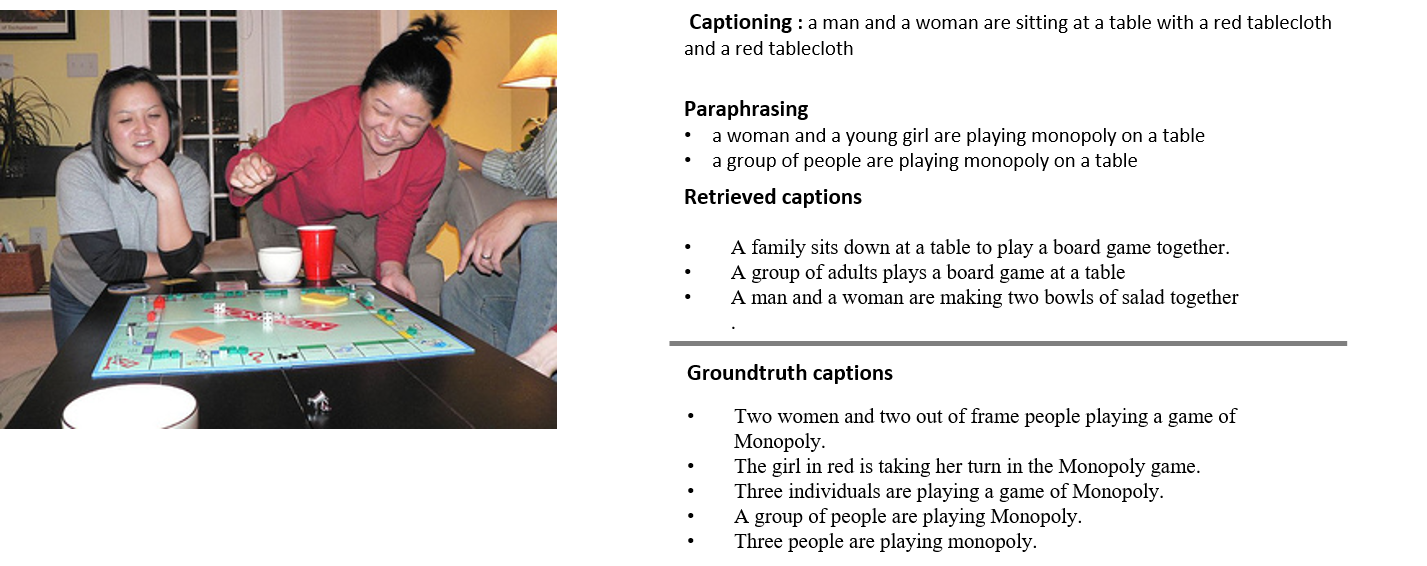}
  \caption{Sample result of STT model on Flickr30K dataset.}
  \label{fig:stt_f30k}
\end{figure}

\subsection{Discussion}
  Figures \ref{fig:stt_coco} and \ref{fig:stt_coco_2} show some sample results of the STT model with attention on MSCOCO. The STT model captures context and relationships between images and captions. 
%   BLEU scores improved on addition of attention in STT model. This can be due to the improved alignment of individual words and image regions for the sentence and image respectively. Since there are all possible combinations of paraphrases during training, the model learnt to align
 All possible permutations of paraphrases were used for training and the model learns to associate image regions to individual words better compared to the model trained without attention. This is evident in Table \ref{im_capt}. The addition of attention improved BLEU \cite{papineni2002bleu} and METEOR \cite{banerjee2005meteor} scores for the captioning task. As the number of samples is higher in MSCOCO \cite{lin2014microsoft}, STT performs better compared to Flickr30K \cite{flickr30k}. Figure \ref{fig:stt_f30k} shows a sample output of STT on Flickr30K \cite{flickr30k} dataset. Although this particular example is a failed retrieval case, the retrieved captions describe the image well. Flickr30K \cite{flickr30k} is a challenging dataset due to high correlation between captions of different images. However, our model exhibits good generalization and diversity in such challenging scenarios.

\section{Conclusion}

We present a novel multi-task model which learns general purpose embeddings for images and captions. This model captures the semantic relationships between vision and language modalities during training and offers an effective way to interpret the intermediate latent representations. We leverage recent attention mechanisms for further performance boosts.  We believe this is the first effort to show sentence paraphrasing in a multi-task setting between vision and language. We evaluate our model on standard benchmark datasets and demonstrate good performance on cross-modal retrieval, image captioning and sentence paraphrasing tasks.

\bibliographystyle{IEEEbib}
\bibliography{egbib}

\end{document}